\documentclass[letterpaper, 10 pt, conference]{ieeeconf}  

\usepackage[left=0.75in,right=0.75in,top=0.75in,bottom=0.75in]{geometry}
\usepackage{multirow}
\usepackage[font=footnotesize]{caption}

\usepackage{balance}

\IEEEoverridecommandlockouts                              

\usepackage{graphics} 
\usepackage{epsfig} 
\usepackage{mathptmx} 
\usepackage{times} 
\usepackage{amsmath} 
\usepackage{amssymb}  
\usepackage[T1]{fontenc} 
\usepackage{dblfloatfix}
\usepackage[noadjust]{cite}
\usepackage{flushend}
\usepackage{graphicx}
\usepackage{booktabs}
\usepackage{textcomp}
\newcommand{\tildehack}{\raisebox{0.5ex}{\texttildelow}}

\usepackage{tikz}

\newcommand\copyrighttext{%
  \footnotesize \textcopyright 2022 IEEE. Personal use of this material is permitted.
  Permission from IEEE must be obtained for all other uses, in any current or future
  media, including reprinting/republishing this material for advertising or promotional
  purposes, creating new collective works, for resale or redistribution to servers or
  lists, or reuse of any copyrighted component of this work in other works.
  }
\newcommand\copyrightnotice{%
\begin{tikzpicture}[remember picture,overlay]
\node[anchor=south,yshift=10pt] at (current page.south) {\fbox{\parbox{\dimexpr\textwidth-\fboxsep-\fboxrule\relax}{\copyrighttext}}};
\end{tikzpicture}%
}

\usepackage{eso-pic}

 \AddToShipoutPicture*{\raisebox{27cm}{\noindent \hspace{0.7cm}This accepted article to IROS 2022 is made available by the authors in compliance with IEEE policy.}}

 \AddToShipoutPicture*{\raisebox{26.6cm}{\noindent \hspace{-14.7cm}Please find the final, published version in IEEE Xplore.}}

\title{\vspace{18pt}\LARGE \bf 
Electroadhesive Clutches for Programmable Shape Morphing of Soft Actuators
}

\author{Gregory M. Campbell$^{1}$, Jessica Yin$^{1}$, Yuyang Song$^{2}$, Umesh Gandhi$^{2}$, Mark Yim$^{1}$, and James Pikul$^{1}$
\thanks{$^{*}$This work was supported in part by National Science Foundation Emerging Frontiers in Research and Innovation (EFRI) award \#1935294 and Graduate Research Fellowship \#202095381. }
\thanks{$^{1}$Department of Mechanical Engineering and Applied Mechanics (MEAM) at University of Pennsylvania, Philadelphia, PA USA, GRASP Lab (email: \{gcampbel, jessyin, yim, pikul\} @ seas.upenn.edu)}%
\thanks{$^{2}$Toyota Research Institute North America (TRINA), Ann Arbor, MI USA}%
}

\begin{document}

\maketitle{}
\copyrightnotice                   
\begin{abstract}

Soft robotic actuators are safe and adaptable devices with inherent compliance, which makes them attractive for manipulating delicate and complex objects. Researchers have integrated stiff materials into soft actuators to increase their force capacity and direct their deformation. However, these embedded materials have largely been pre-prescribed and static, which constrains the actuators to a predetermined range of motion. In this work, electroadhesive (EA) clutches integrated on a single-chamber soft pneumatic actuator (SPA) provide local programmable stiffness modulation to control the actuator deformation. We show that activating different clutch patterns inflates a silicone membrane into pyramidal, round, and plateau shapes. Curvatures from these shapes are combined during actuation to apply forces on both a 3.7 g and 820 g object along five different degrees of freedom (DoF). The actuator workspace is up to \hbox{12 mm} for light objects. Clutch deactivation, which results in local elastomeric expansion, rapidly applies forces up to \hbox{3.2 N} to an object resting on the surface and launches a 3.7 g object in controlled directions. The actuator also rotates a heavier, 820 g, object by 5 degrees and rapidly restores it to horizontal alignment after clutch deactivation. This actuator is fully powered by a 5 V battery, AA battery, DC-DC transformer, and 4.5 V (63 g) DC air pump. These results demonstrate a first step towards realizing a soft actuator with high DoF shape change that preserves the inherent benefits of pneumatic actuation while gaining the electrical controllability and strength of EA clutches. We envision such a system supplying human contact forces in the form of a low-profile sit-to-stand assistance device, bed-ridden patient manipulator, or other ergonomic mechanism.

\end{abstract}

\section{INTRODUCTION}

Soft robotic actuators use elastic deformation to produce smooth, continuous motions \cite{robinson1999continuum} and conform to delicate objects \cite{pettersson2010design} while remaining resilient to adverse environmental conditions or critical loading \cite{tolleymichael2014resilient}. Soft materials are generally inexpensive, light, and easily compactable for storage or transportation. These properties make soft actuators useful across a variety of applications including locomotion, gripping, manipulation, and haptic response \cite{laschi2016soft,yin2020wearable}. 

\begin{figure}[!t]
      \centering
    \includegraphics[width=\linewidth]{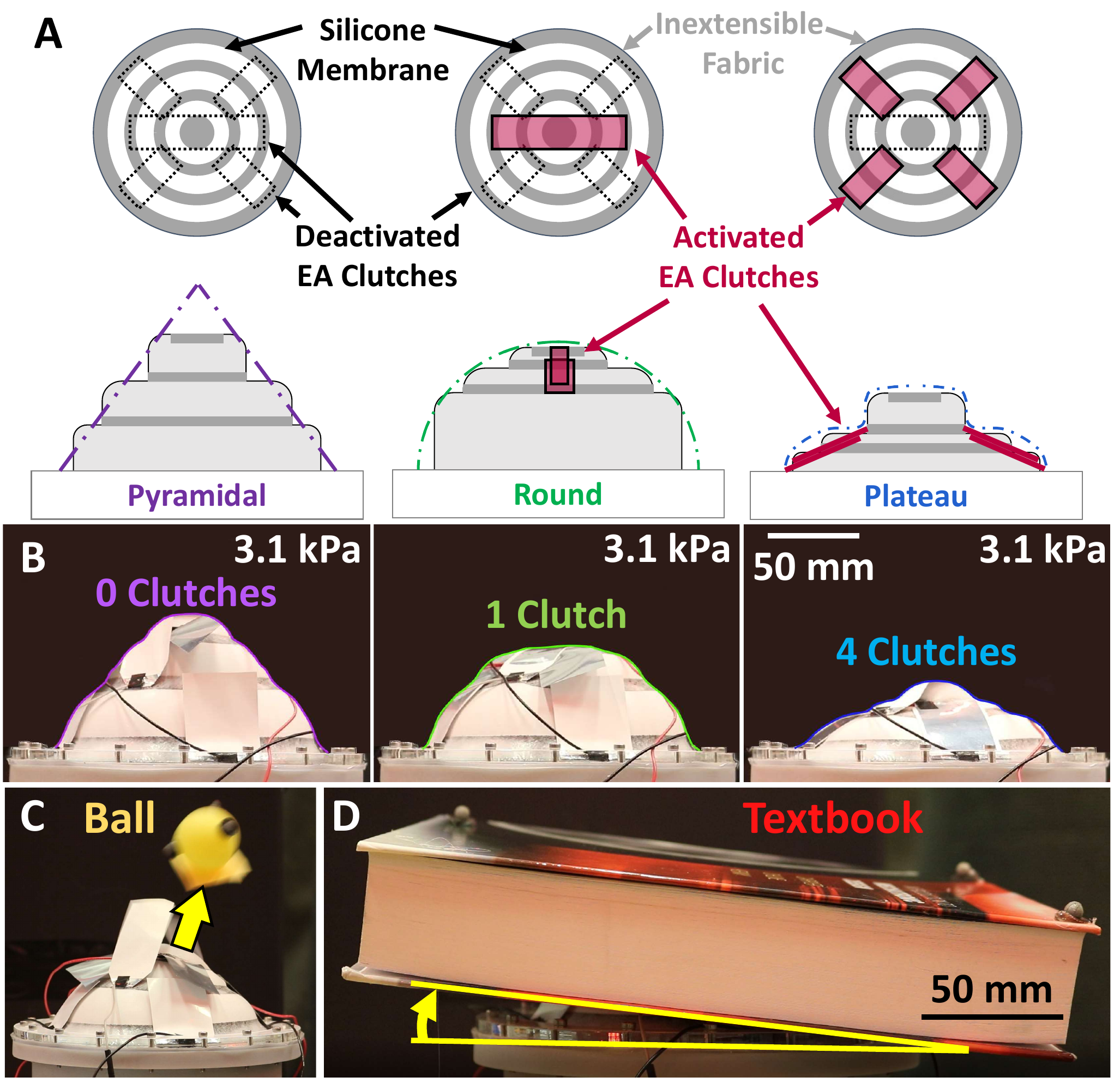}
      \caption{A. Actuator expansion formed into three different shapes; shape chosen by clutch activation. Top-view presented over side-view. B. Expansion of soft, elastomeric actuator to 3.1 kPa under three different clutch activation configurations. C. Soft actuator manipulation of 3.7 g ball. D. Manipulation of 820 g textbook.}
      \label{Figure1}
      \vspace{-6mm}
  \end{figure}
  
Most soft pneumatic actuators are constrained to deform along a single determined range of motion due to the static nature of their material composition. These single-configuration soft pneumatic actuators (SPA) are fundamentally limited in their ability to controllably adapt to new shapes, actuation modes, and forces. SPAs are typically made from pressurized chambers or "pneu-nets" in elastomers, which control the scale or speed of material expansion \cite{shepherd2011multigait,mosadegh2014pneumatic}. SPAs generally use passive inextensible materials, such as fabric \cite{pikul2017stretchable} or fibers \cite{sholl2021controlling},
to restrict expansion towards a single desired range of motion.  If an alternative range of motion is desired, an entirely new actuator must be manufactured. Achieving general 3-dimensional motion control with this type of SPA necessitates the use of multiple separate chambers working in tandem \cite{shepherd2011multigait,balak2020multi}. These SPAs therefore require multiple pressure sources and rely on rigid peripherals for electrical controllability across multiple degrees of freedom (DoF). An alternative approach is to actively program the stiffness of materials in soft actuators and, therefore, allow the range of motion to be modulated in real time with a single source of pressurized fluid.

Active stiffness modulation of soft materials has been explored with technologies such as granular jamming \cite{steltz2009jsel}
and joule heating \cite{firouzeh2015soft}, 
both of which rely on long transition times (multiple seconds). Faster, electrostatic solutions have also been realized \cite{levine2021materials}, including dielectric elastomer actuators (DEA), which align opposite charges around a dielectric elastomer to reduce its thickness and increase its area. Electrostatic chucking \cite{imamura2017variable} stacks many layers of DEAs to increase stiffness, while DEA membranes \cite{2014ZouActive} 
use the DEA to decrease their already low stiffness. Electroadhesive (EA) clutches are an attractive alternative because they have been shown to modulate stiffness in tens of milliseconds while applying stresses over 8 N/$cm^2$ at 400V \cite{hinchet2020high,hinchet2018dextres,diller2018effects}. EA clutches apply opposite electric charge on two parallel pads, which are then adhered to each other and prevent relative motion. EA pads have been used to modulate stiffness for restricting human motion \cite{diller2016lightweight, hinchet2020high} and in soft robots to assist with grasping \cite{guo2018adaptive} 
or modulate physical connections \cite{germann2012active}. EA clutches present an opportunity for soft actuation, as they are light weight, flexible yet inextensible, and require no off board pumps or motors. 

This work demonstrates the potential of EA clutches to provide stiffness modulation and shape control for a single-chamber soft pneumatic actuator. By attaching clutches to a membrane and alternating clutch activation, we demonstrate the ability to vary stiffness and inflate into multiple shapes and curvatures. 
These curvatures are primitives for a broader set of complex shapes, as previously demonstrated for camouflage applications \cite{pikul2017stretchable}. We demonstrate the versatility of this electrical stiffness control in two modes of actuation along five DoF: expansion-driven manipulation of a \hbox{3.7 g} object and pneumatic manipulation of a \hbox{820 g} object. We use manipulation of a \hbox{3.7 g} object to characterize the actuator workspace, up to \hbox{12 mm} along the five DoF, and to assess the magnitude and consistency of force applied by local membrane expansion caused by deactivating a clutch, up to \hbox{3.2 N}. The actuator also tilts a \hbox{820 g} object by \hbox{5 degrees} via pneumatic inflation with clutch activation and rapidly repositions it via clutch deactivation. Using clutch deactivation instead of pressure increase to rapidly apply force allows this actuator to function untethered, with a small (4.5 V, 63 g) air pump supplying pressure and two batteries (5V, AA) as energy sources.

This work represents the first usage of EA clutches for soft actuator stiffness modulation and the first instance of real-time electrically-controllable soft membrane stiffness modulation for pneumatic force application. The actuator presented is mobile and versatile, lightweight and inexpensive, and provides both speed and strength. 
This work presents critical progress towards scalable, real-time variable, and broadly applicable soft actuators that achieve high strain and electrically-controllable actuation. 

\begin{figure}[th!]
\centering
\includegraphics[width=0.8\linewidth]{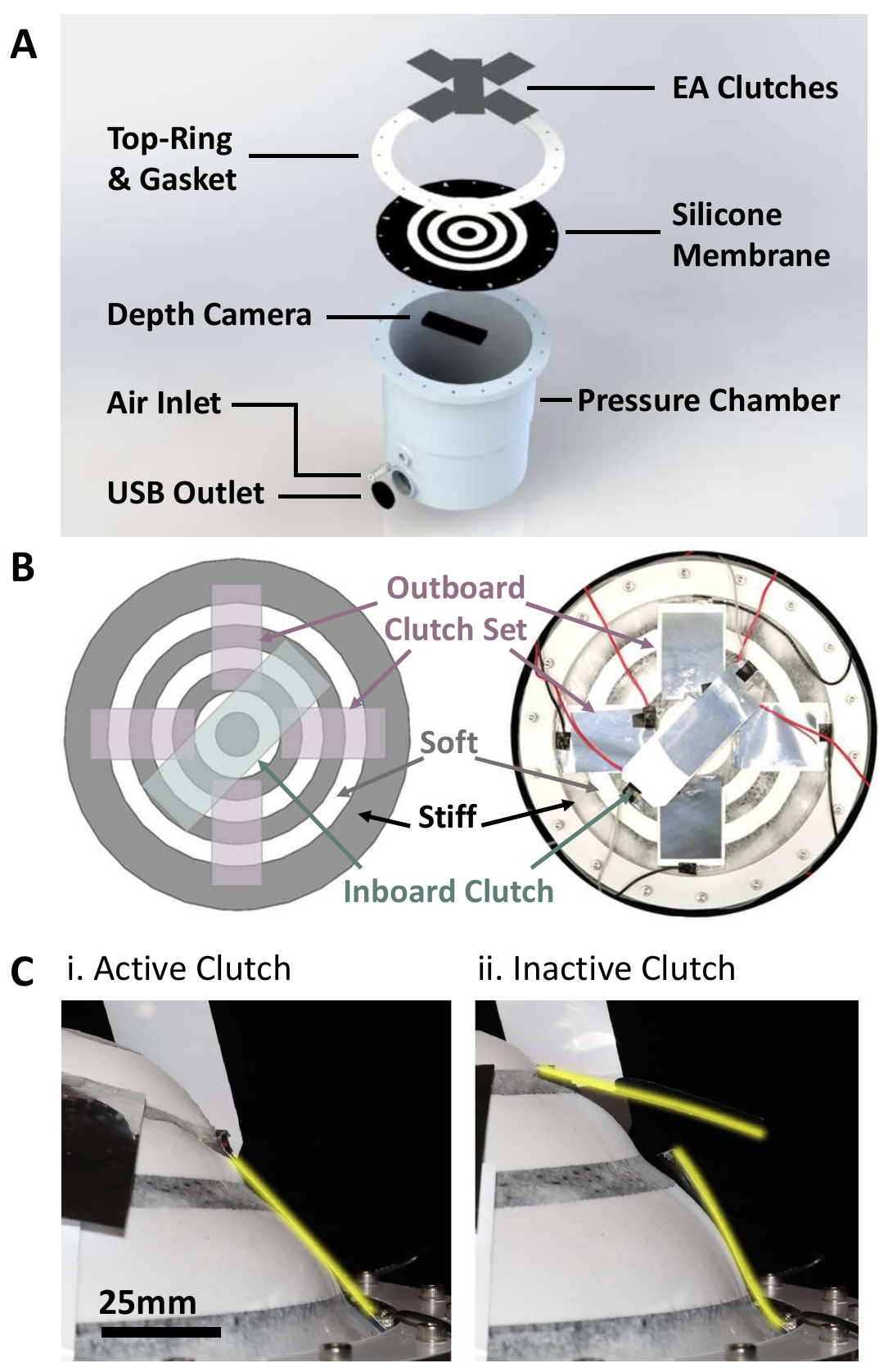}
\caption{System Overview. A. Overview of actuator test system with labels. B. Model of silicone membrane with clutch locations superimposed. Model includes stiff silicone reinforced with fabric, soft unsupported silicone, four outboard clutches, and one inboard clutch. C. Clutch is highlighted with yellow. i) Outboard clutch is activated and restricts membrane expansion. ii) Outboard clutch is deactivated and membrane is free to expand.}
\label{Membrane Layout}
\vspace{-6mm}
\end{figure}

\section{System Development \& Fabrication}

Our soft membrane testing system (Figure \ref{Membrane Layout}A) inspired by \cite{alspach2019soft} consists of three main components: A) a pressurized chamber, B) the elastomeric membrane with EA clutches, and C) sensors and control electronics.


\subsection{Soft Membrane Test System - Chamber}
The pressurized chamber is assembled from two 3D-printed pieces. 
Both pieces are glued together using a resin epoxy adhesive. 
The chamber has two ports: an air inlet via a push-to-connect tube fitting and a dual USB 3.0 port for sensor data output. Both ports are sealed with silicone gaskets and resin epoxy adhesive. The elastomeric membrane is clamped down at the top of the chamber with a silicone gasket, laser-cut acrylic ring, and screws. The inner diameter of the top ring restricts the expanding area of the actuator to a diameter of \hbox{150 mm}. This diameter was governed by the size of the 3d-printer print-bed, but can be scaled up or down with the same clutch pattern and electrical controllability.\\

\vspace{-3mm}

\subsection{Elastomeric Membrane and Electroadhesive (EA) clutches}

To make the membrane, we laser-cut Soft N' Shear fabric stabilizer and place it in Ecoflex 00-30 during curing to provide areas of high stiffness. We then connect EA clutches to these fabric-reinforced areas via Sil-Poxy silicone adhesive. 

We designed the Soft N' Shear stiff regions 
based off the zero Gaussian curvature layout from \cite{pikul2017stretchable}. The concentric rings create a pyramidal shape upon inflation
We space the rings such that there is a large enough stiffened region for adhering the EA clutch plates. We create three concentric regions of unstiffened silicone (Figure \ref{Membrane Layout}B) and control the activation of EA clutches to restrict the expansion of some subset of these three soft regions (Figure \ref{Membrane Layout}C). We can activate multiple clutches in parallel, which allows for pre-determined patterns to activate with a single signal. For the shape change configuration (Figure \ref{Membrane Layout}B) the red, outboard, clutches are activated by one signal, while the green, inboard, clutch is activated by another. Clutch and fabric positioning can be altered on subsequent membranes for separate types of applications.

The EA clutches use Dupont Luxprint 8153 as a dielectric. The parameters of Luxprint clutches are well characterized \cite{diller2018effects} and provide sufficiently high forces at hundreds of volts relative to other dielectric options \cite{hinchet2019high,hinchet2020high}. Following the fabrication process specified by Diller et al. \cite{diller2018effects}, we applied a nominally 50 $\mu m$ layer (actual thickness varies from 40-50 $\mu m$) of Luxprint to nominally 50 $\mu m$ thick aluminum-sputtered biaxially-oriented polyethylene terephthalate (BOPET).
 We cut the resulting sheets into clutch plates and attach wires with MG Silver Epoxy 8331. Electrical tape provides electrical insulation over wire leads and prevents delamination of the Epoxy. Clutches are activated at a DC voltage of approximately \hbox{420 V}. 

\begin{figure}[b!]
      \centering
      \includegraphics[width=0.5\linewidth]{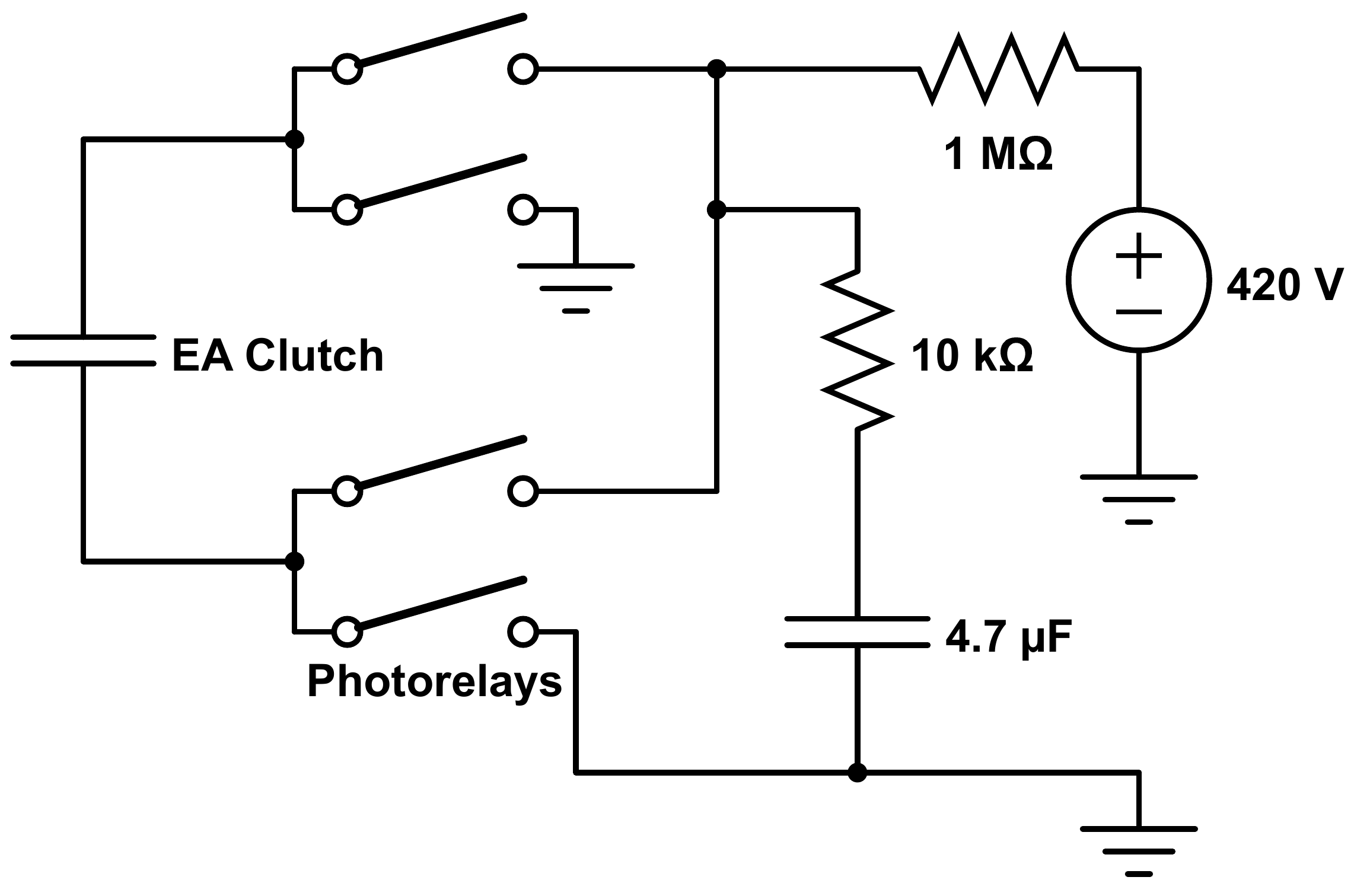}
      \caption{Circuit Diagram. Clutch circuit diagram, building upon \cite{diller2018effects}.}
      \label{Circuit}
\end{figure}

\subsection{Electronics}
The pressure chamber houses an air pressure sensor (Qwiic MicroPressure Sensor; Sparkfun), an ESP32 microcontroller, and a time-of-flight depth camera (Picoflexx; PMD). The air pressure sensor monitors the internal air pressure and connects to the microcontroller to output data through the USB port. The depth camera captures membrane deformation data by measuring the displacements of the membrane from inside the chamber and passes data through the USB port. 

External to the pressure chamber, a ZR370-02PM \hbox{4.5 V} DC air pump and vacuum (dimensions: 58x27x27 mm, mass: 63 g) powered by a 5 V battery applies air pressure. The electronic clutch control circuit (Figure \ref{Circuit}) includes TLP222G-2 photorelays which switch to bring one clutch plate to the high (\tildehack420 V) voltage \cite{diller2018effects}. $4.7 \mu F$, 400 V capacitors supply  additional current in parallel with the high voltage supply during activation transitions. Using two relays allows  each clutch to switch polarity. Switching polarity  between tests counteracts space-charge effects, enabling reliable deactivation. This circuitry was scaled up to allow for the simultaneous or subsequent activation of multiple clutches. An ESP32 microcontroller controls inflation and clutch (de)activation with inputs from a laptop computer. Similarly, the laptop records pressure data and depth images. An EMCO F101CT DC/DC converter connected to a AA battery supplies power for the clutches.



 \begin{figure}[tb!]
      \centering
      \includegraphics[width=\linewidth]{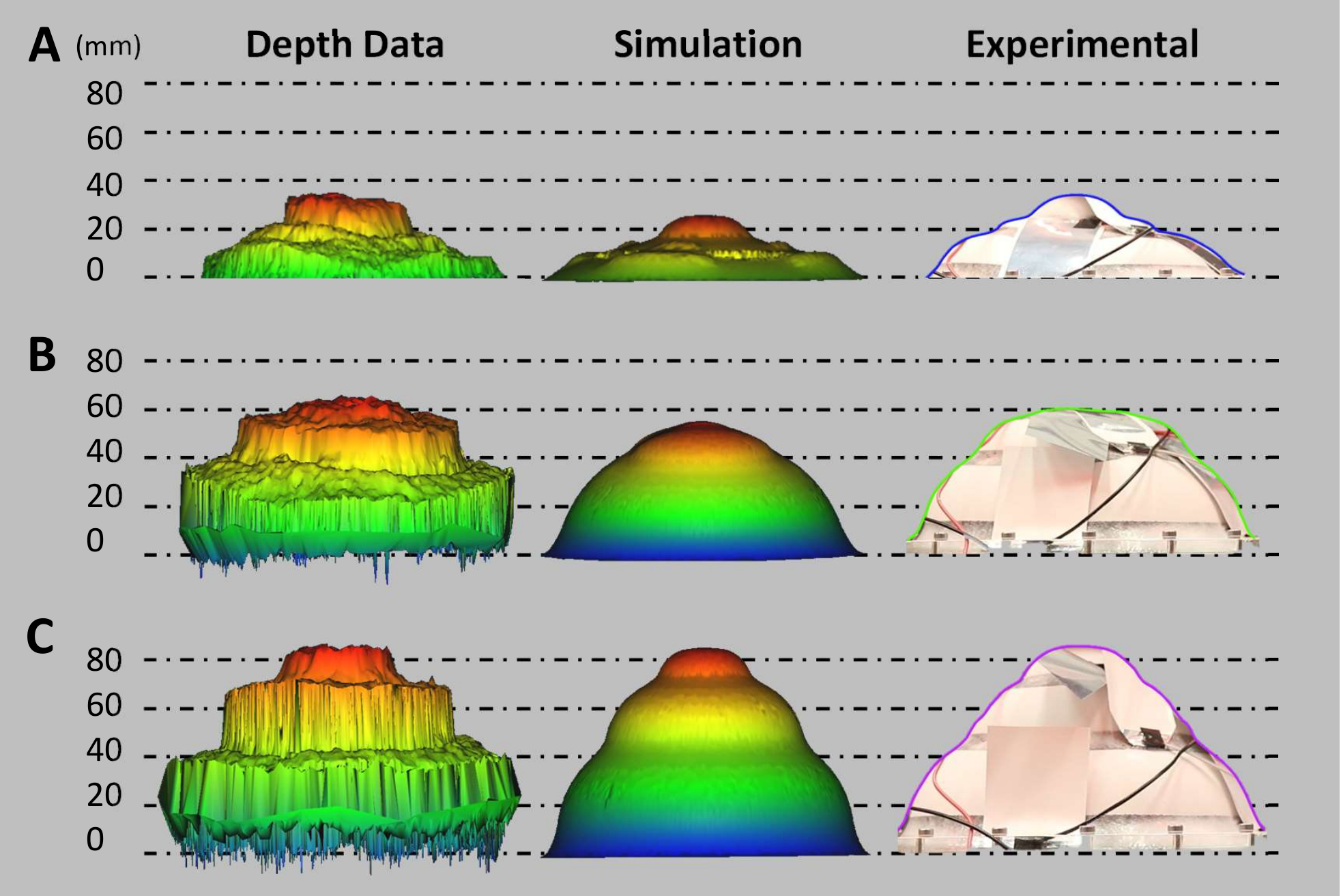}
      \caption{Depth camera data (left), simulation (center), and experimental system (right) for actuator shape comparison at 3.1 kPa. A. Plateau shape. B. Round shape. C. Pyramidal shape.}
      \label{FEA_Results}
      \vspace{-4mm}
      
\end{figure}
\section{Results \& Discussion}
  \begin{figure*}[t!]
      \centering
      \includegraphics[width=\textwidth]{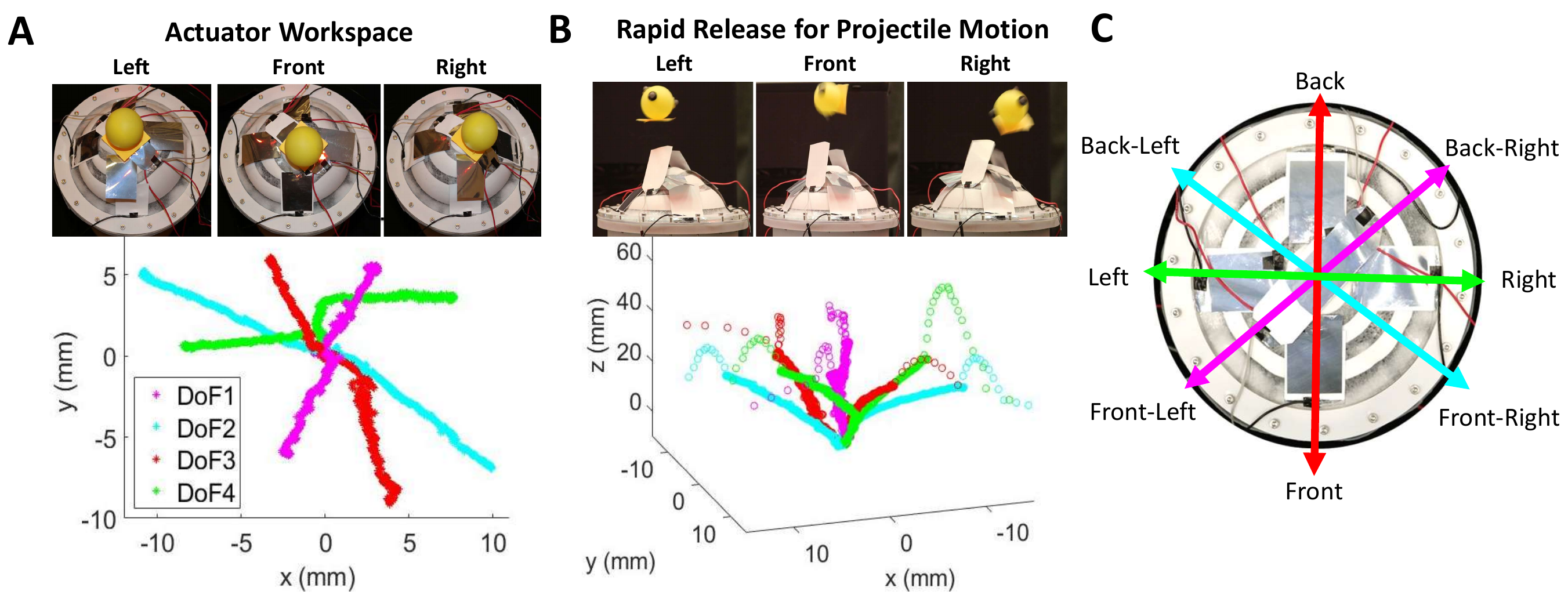}
      \caption{Mode 1 Actuator Characterization. A. Plot of actuator workspace along 4 DoF, looking at the actuator from a top-down view. Each DoF is accessed by activating a different set of clutches. Photos of corresponding positioning of 40 mm diameter ball. B. Plot that tracks position of ball once it has been placed on the actuator. The actuator lifts it to the specified direction and then releases the inboard clutch to rapidly apply force to the ball in the desired direction. Photos of corresponding projectile motion for 40 mm diameter ball. C. Labeled photo of actuator with directions and DoFs.}
      \label{Actuator-Characterization}
      \vspace{-4mm}
  \end{figure*}
\subsection{Finite element modeling}

We use the finite element (FE)  package ABAQUS 2020/Standard \cite{Abaqus} to understand how the inflatable structure behaves under different clutch constraints. In this analysis, we model the silicone inflatable structure as a circular shell discretized into S4R element types. 
The thickness of the silicone shell and the attached clutches match the experimental setup at 1 mm and 0.2 mm, respectively. 
  
Pneumatic pressure is applied normal of the top surface of the membrane, while the edge of the structure is fixed with a zero-displacement constraint in all 6 DoF. Based on the membrane design, we assigned each portion of the ring to either Ecoflex 00-30 or Soft N' Shear fabric stabilizer (high stiffness). 

We model the Ecoflex's hyperelastic behavior using a 3-term Ogden model 
\cite{softrobotictools}, Soft N' Shear areas using elastic material with a modulus of 8 MPa, and the clutch, including the two individual clutch plates and adhesive connections, using elastic material with a modulus of 100 MPa.
The clutch is attached with perfect bonding to the membrane over the entirety of the regions marked in Figure \ref{Membrane Layout}B. This differs from the experimental system, in which the clutches  bond to the membrane only at the ends.
The simulation runs dynamically, without mass scaling. The output displacement contour provides position values.

Our modeling predicts that thin clutches can provide significant change in  membrane inflation shape when inflated to 3.1 kPa (gauge pressure), as seen in the center column of Figure \ref{FEA_Results}. Four thin clutches are positioned in the outboard positions for Figure \ref{FEA_Results}A, one clutch is positioned at the inboard position for Figure \ref{FEA_Results}B, and the stabilized silicone is simulated without any clutches in Figure \ref{FEA_Results}C. The maximum height the membrane reaches for each simulation respectively is: \hbox{26.5 mm}, 57.1 mm, and 86.3 mm.

\begin{table}[t!]
\centering
\renewcommand{\arraystretch}{1.4} 
\caption{\normalsize Forces, Standard Deviations, and Average Directional Consistencies for Each DoF and Direction.}
\begin{tabular}{ccccc}
\hline
\textbf{DoF}           & \textbf{Direction} & \textbf{Force (N)} & \textbf{Std (N)} & \textbf{Consistency (\%)} \\ 
\hline
\multirow{2}{*}{DoF 1} & Back-Right          & 1.2   & 0.21  & 99.3          \\ 
                      & Front-Left          & 1.4    & 0.28  & 99.5            \\ 
                      \hline
\multirow{2}{*}{DoF 2} & Front-Right         & 1.1    & 0.41 & 99.8           \\ 
                      & Back-Left           & 1.4    & 0.19  & 99.4            \\ 
                      \hline
\multirow{2}{*}{DoF 3} & Front               & 2.4   & 0.35   & 99.4             \\ 
                      & Back                & 2.7    & 0.72     & 98.1       \\ 
                      \hline
\multirow{2}{*}{DoF 4} & Left                & 3.2      & 0.57      & 99.6    \\ 
                      & Right               & 2.2      & 0.71       & 99.3   \\ 
                      \hline
DoF 5                  & Up                  & 2.8      & 1.1       & 99.8   \\ 
\hline
\end{tabular}
\label{tab:force_table}
\vspace{-4mm}
\end{table}

\subsection{Experimental Results}

The left column of Figure \ref{FEA_Results} shows the time-of-flight depth data of the membrane as we inflate it to \hbox{\tildehack3.1 kPa} with three different clutch configurations. Prior to and throughout inflation to the plateau shape (Figure \ref{FEA_Results}A), we apply the voltage difference of \tildehack420 V across the four outboard clutches (red rectangles in Figure \ref{Membrane Layout}B), restricting the expansion of the two outboard rings of unsupported silicone. The resulting expansion therefore occurs primarily at the innermost ring of silicone, leading to a height of 32.7 mm. Prior to and throughout inflating the membrane to the round shape (\hbox{Figure \ref{FEA_Results}B}), we activate only the interior clutch. When fully inflated, the membrane reaches a height of 66.7 mm. We activate no clutches for the pyramidal shape (Figure \ref{FEA_Results}C), leading to an inflation height of 86.3 mm.

To compare the experimental and simulation results, we align the point clouds using the Iterative Closest Point algorithm and calculated the average root-mean squared error (RMSE) for each shape with CloudCompare \cite{girardeau2016cloudcompare}. We use standard point cloud pre-processing techniques on the experimental data, including statistical outlier removal and the k-Nearest Neighbor noise filter. Average RMSE for the plateau, round, and pyramidal shapes respectively were \hbox{4.4 mm}, 6.1 mm, and 6.3 mm.

\subsection{Actuation}

\begin{figure}[t!]
      \centering
      \includegraphics[width=.7\linewidth]{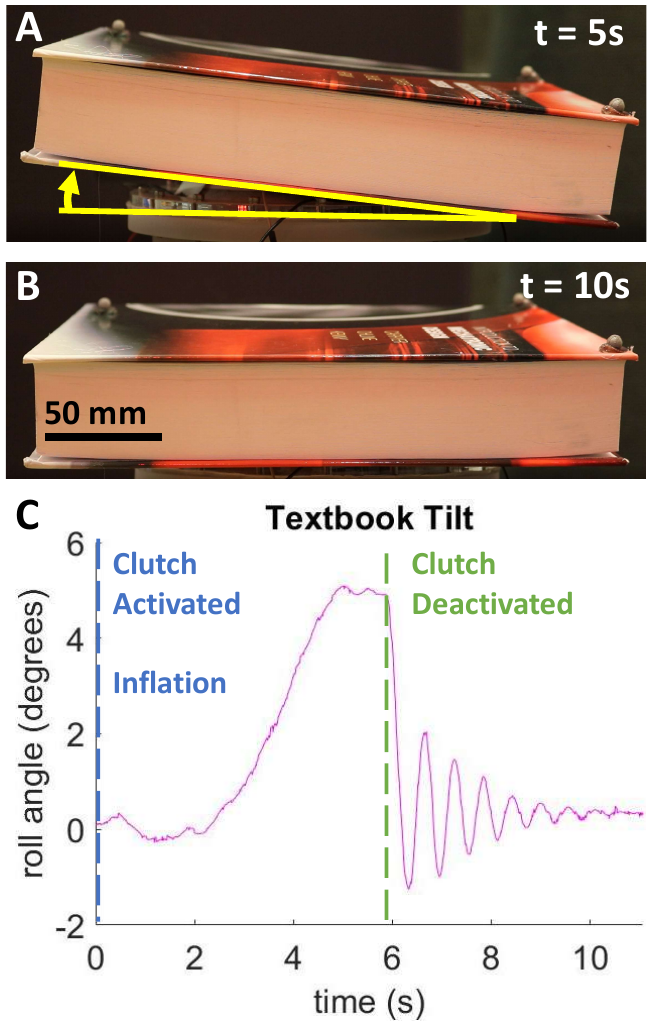}
      \caption{Mode 2 Manipulation. A. Textbook lifted and tilted by pneumatic inflation to 3.1 kPa during right clutch activation. B. Textbook returned from tilt by clutch deactivation. C. Vicon data for textbook roll angle response to inflation under clutch activation and to clutch deactivation.}
      \label{Book}
      \vspace{-6mm}
\end{figure}

We leverage shape-changing characteristics of the membrane into two separate modes of actuation, exemplified in Figure \ref{Actuator-Characterization} and Figure \ref{Book} respectively. Mode 1 emphasizes the fast responsiveness of the actuator enabled by the electric signal control for EA clutches, while Mode 2 emphasizes the high force density of pneumatic actuation, while still providing active control via the EA clutches. Between expansions, the membrane is deflated by opening the pressure chamber to the ambient air and clutches are re-aligned by hand.

\vspace{3mm}
\subsubsection{Mode 1}

Mode 1 allows for the manipulation of light objects that have minimal effect on the shape of the membrane. We activate clutches relevant to the desired motion at the start of inflation, and after inflation is complete, we deactivate the inboard clutch to apply a rapid force in the desired direction.

We evaluate the actuator's workspace by altering clutch configurations in Mode 1. We induce motion along cardinal directions by activating a single outboard clutch and the inboard clutch, and ordinal directions by activating two adjacent outboard clutches and the inboard clutch. We position a light payload (ping pong ball, Vicon markers, and paper for stability - mass: \hbox{3.7 g}) along one of five different DoF.
We monitor ball position in three dimensions via Vicon cameras while inflating the membrane from 0 to \hbox{1.7 kPa} in separate trials for each cardinal and ordinal direction. The actuator displaces the ball between 6 and \hbox{12 mm} along each degree of freedom as displayed in Figure \ref{Actuator-Characterization}A, this represents up to \hbox{16 \%} of the 75 mm radius. The fifth and final degree of freedom, `Up', results from no clutches active and is in the \textit{z} plane, perpendicular to the plot in Figure \ref{Actuator-Characterization}A. 

After the membrane is inflated, we deactivate the inboard clutch, which causes the locally restrained membrane to rapidly "soften." This leads to a rapid expansion of the elastomer, which applies a force on the object in a direction corresponding to the actuator shape. Figure \ref{Actuator-Characterization}B shows subsequent projectile motions that appear as parabolas beginning at the edge of the actuator workspace. Positions are plotted at 0.01 second intervals. Sparser data indicates faster motion induced by clutch deactivation. 

We characterized force output separately from workspace. In this characterization, the membrane was inflated to \hbox{2.8 kPa} for cardinal directions (and `Up') and \hbox{1.7 kPa} for ordinal directions. Outboard clutches were subject to failure at pressures higher than this. Data from trials involving clutch failure is not included. We then deactivated the inboard clutch, causing the force response discussed for \hbox{Figure \ref{Actuator-Characterization}B}. We calculated force based on the motion of the ball as follows: we calculated the components of acceleration for the ball in each direction (\textit{x},\textit{y},\textit{z}) from the second derivative of its position with respect to time and identified the instantaneous acceleration most relevant to the actuator force. We adjusted the  z component of acceleration by summing it with the acceleration due to gravity. We calculated magnitude of force as the product of the mass of the ball and the $L^2$-norm of the three components of acceleration. We conducted five trials for each direction\footnote{Only three trials were conducted in the `Up' direction}, and report the averages and standard deviations of force magnitudes in Table \ref{tab:force_table}. 

We also analyzed the direction of force by calculating the unit vector associated with the average force applied in each DoF. We took the dot product of this average unit vector and the unit vector associated with each trial to solve for the directional consistency of that trial on a scale of -1 (complete opposite direction) to 1 (identical direction). We report the average of these scalar values for the five trials as a percentage in Table \ref{tab:force_table}, where a dot product of 1 represents 100\% consistency. In every trial, the primary component of force is in the upward direction.

\vspace{3mm}
\subsubsection{Mode 2}

Mode 2 allows for the manipulation of heavier objects that alter the shape profile of the membrane. 
These types of object cannot be instantaneously accelerated away from the membrane as in Mode 1. We activate the desired outboard clutch(es) prior to inflation, which causes a tilt during the object lift. Subsequent to inflation, deactivating outboard clutches provides a rapid actuator shape change. 

We inflate the actuator to 3.1 kPa with the right clutch active to bring the roll angle of the textbook (textbook with Vicon markers - mass: 820 g) to 5 degrees (Figure \ref{Book}A). We subsequently deactivate the EA clutch, allowing the mass of the textbook and material properies of the inflated membrane to return the roll angle to approximately zero (Figure \ref{Book}B). Vicon data displays the textbook's roll angle versus time in Figure \ref{Book}C, with inflation leading to a gradual increase in roll angle and clutch deactivation leading to a rapid decline in roll angle and an underdamped return to equilibrium.

\subsection{Discussion of Results}

We validate all three simulated shapes with high accuracy using the proposed experimental setup. The plateau shape represents local areas of negative Gaussian curvature and the round shape represents positive Gaussian curvature. The combinations of these different curvatures allow a range of different positions within the actuator workspace, which reaches up to \hbox{12 mm} along various degrees of freedom. Different combinations of membrane and clutch layouts could be fabricated to reach myriad different shapes and motions.

Discrepancies between experimental and simulation results occur primarily at clutch-restricted regions. These discrepancies could be due to the model's assumption that the clutch plates are fully overlapped at activation, and that no slipping occurs between activated clutch plates at any point during inflation. Both these assumptions can change in reality. Nonlinear behavior of the Soft N' Shear stiffening agent is ignored in the simulation, which could also contribute to discrepancies. The close correlation between the modeled and experimental results shows promise for the use of FEM with EA clutch placement on a soft membrane.

The ability to induce projectile motion of the 3.7 g ball during Mode 1 manipulation shows that this actuator is capable of rapid directional force application at the edges of its workspace. This type of explosive force could be useful in a lightweight mobile robot, particularly for a jumping motion. The inconsistent ranges of motion and force applications across different degrees of freedom are likely due to imprecisions in membrane manufacturing (we aligned clutches by hand). 

Without any reconfiguration, the actuator can then actuate in Mode 2 to position the 820 g textbook pneumatically and rapidly reposition it with clutch deactivation. Figure \ref{Book} represents a single trial, and Mode 2 was significantly less consistent between trials, as small offsets in the  position of the textbook's center of mass greatly altered the trajectory. Comparably sized Ecoflex pneumatic membranes have been shown to lift up to 5.84 kg \cite{sholl2021controlling}, but we did not lift more than 820 g for this study. This lifting capacity could be increased with a larger actuator or by using more actuators in parallel. While this system won't reach the force output or precision of a rigid platform, its natural compliance can provide safety and comfort in the application of human contact forces. 

\section{CONCLUSIONS \& FUTURE WORK}

This work demonstrates EA clutches as a viable means of stiffness modulation for inflatable soft actuators. Furthermore, it shows that we can use clutches to alter the inflation of a single soft membrane to accurately recreate three target shapes and manipulate both light and heavy objects along five degrees of freedom while supplying air only from a small, low-voltage, DC pump. A light object manipulation workspace is defined for the actuator, and forces are characterized at the edges of this workspace. This actuator characterization and shape changing demonstration, along with the ability to predict and design for shapes with modeling, lays the groundwork for more advanced pneumatically actuated soft robots. This technology, applied locally on a larger membrane or on a series of membranes working in parallel, could provide a mechanism for powerful, soft manipulation.

There are a number of remaining challenges to be overcome in future work. The actuator needs to be made more robust for real-world use, as failure frequently occurs at clutch lead connections and membrane-clutch adhesive interfaces. EA clutch reliability and robustness needs to be addressed to eliminate the need for careful alignment prior to clutch activation and regular cleaning of clutch plates between trials. If clutch reliability was addressed, this actuator could be controlled in more arbitrary degrees of freedom via time-dependent clutch activation during inflation. This actuator could also benefit from more precise control, specifically closed-loop clutch deactivation based on Vicon camera or internal depth camera data, which would allow for greater consistency during \hbox{Mode 2} manipulation. Refined \hbox{Mode 2} manipulation could provide direct human contact forces in a sit-to-stand assistance device,  bed-ridden patient manipulator, or other ergonomic mechanism.







\bibliographystyle{IEEEtran}
\bibliography{bibliography}
\balance

\end{document}